\documentclass[11pt]{article}

\usepackage[]{acl22}

\usepackage{times}
\usepackage{latexsym}

\usepackage[T1]{fontenc}

\usepackage[utf8]{inputenc}

\usepackage{microtype}

\usepackage{amsmath}
\usepackage{amssymb}
\usepackage[ruled, lined]{algorithm2e}
\usepackage{comment}
\SetAlFnt{\small}
\usepackage{helvet} 
\usepackage{courier}  
\usepackage{graphicx} 
\usepackage{multirow} 
\usepackage{natbib}  
\usepackage{caption} 
\usepackage{xcolor}
\usepackage{epsfig}

\title{Modularized Transfer Learning with Multiple Knowledge Graphs \\ for Zero-shot Commonsense Reasoning}

\author{Yu Jin Kim\textsuperscript{\rm 1}, Beong-woo Kwak\textsuperscript{\rm 1}, Youngwook Kim\textsuperscript{\rm 1}, Reinald Kim Amplayo\textsuperscript{\rm 2}\thanks{~~Work done while at the University of Edinburgh.}\\\textbf{Seung-won Hwang}\textsuperscript{\rm 3} and \textbf{Jinyoung Yeo}\textsuperscript{\rm 1}\thanks{~~Corresponding author}\\
  \textsuperscript{\rm 1}Yonsei University, Seoul, Korea 
  \textsuperscript{\rm 2}Google Research, London, UK\\
  \textsuperscript{\rm 3}Seoul National University, Seoul, Korea\\
  \texttt{\{yujin000731, beongwoo.kwak, youngwook, jinyeo\}@yonsei.ac.kr}\\
  \texttt{reinald@google.com, seungwonh@snu.ac.kr} \\}

\newcommand{\eg}{{\it e.g.}}%
\newcommand{\ie}{{\it i.e.}}%
\newcommand{\etc}{{\it etc}}%
\newcommand{\calD}{\mbox{${\cal D}$}}

\newcommand{\calL}{\mbox{${\cal L}$}}

\begin{document}
\maketitle

\begin{abstract}
Commonsense reasoning systems should be able to generalize to diverse reasoning cases. However, most state-of-the-art approaches depend on expensive data annotations and overfit to a specific benchmark without learning how to perform general semantic reasoning. To overcome these drawbacks, zero-shot QA systems have shown promise as a robust learning scheme by transforming a commonsense knowledge graph (KG) into synthetic QA-form samples for model training. Considering the increasing type of different commonsense KGs, this paper aims to extend the zero-shot transfer learning scenario into multiple-source settings, where different KGs can be utilized synergetically. Towards this goal, we propose to mitigate the loss of knowledge from the interference among the different knowledge sources, by developing a modular variant of the knowledge aggregation as a new zero-shot commonsense reasoning framework. Results on five commonsense reasoning benchmarks demonstrate the efficacy of our framework, improving the performance with multiple KGs.
\end{abstract}
\section{Introduction}\label{sec:intro}
The ability to understand natural language through commonsense reasoning is one of the core focuses in the field of natural language processing. To measure and study the different aspects of commonsense reasoning, several datasets are developed, such as SocialIQA~\citep{sap2019socialiqa}, CommonsenseQA~\citep{talmor2018commonsenseqa}, and PhysicalIQA~\citep{bisk2020piqa}, each requiring different type of commonsense knowledge (\eg, social, taxonomic, causal, declarative, \etc) to select the correct answer. While large-scale neural systems~\citep{devlin2018bert,yang2019xlnet,liu2019roberta} have shown human-level accuracy on these benchmarks, recent studies~\citep{mitra2019exploring} also criticize that these models solve individual datasets, rather than learning how to perform general semantic reasoning. To this end, \citet{ma2020knowledgedriven} suggested zero-shot evaluation as a genuine measure for the reasoning capability of the machine.

Inspired by this new metric, in this work, we focus on building unsupervised zero-shot multiple-choice QA systems. That is, we target an arbitrary commonsense reasoning task where conventional approaches (that rely heavily on task-specific supervision) are not applicable to such zero-shot learning scenarios. To learn QA models without expensive annotation efforts, recent works~\citep{ma2020knowledgedriven, banerjee2020self, malaviya2020commonsense} propose to generate a synthetic QA dataset using a commonsense KG such as \texttt{ATOMIC}~\citep{sap2019atomic} and \texttt{ConceptNet}~\citep{speer2017conceptnet}. Such an approach mostly focuses only on one specific type of reasoning relations (\eg, if-then relation, or declarative relation), neglecting the fact that real-world QA systems require simultaneously considering different types of reasoning abilities (\eg, declarative and social, or causal and physical reasoning; \citealp{ilievski2021dimensions, chang2021incorporating}). 

To consider different types of reasoning, this paper extends ideas from the aforementioned zero-shot learning to the \emph{multi-source} case such that it benefits from different types of commonsense knowledge on individual KGs. For example, \texttt{ATOMIC}~\citep{sap2019atomic} focuses on social commonsense while \texttt{ConceptNet}~\citep{speer2017conceptnet} contains conceptual knowledge. A practical approach is multi-task learning (MTL; \citealp{caruana1997multitask, liu2019multi}), which learns a shared encoder for different synthetic QA datasets from multiple KGs. Despite its effectiveness, MTL scheme suffers from interference among different KGs, which results in forgetting previously learned knowledge when trained on new KG which has different kinds of knowledge~\citep{pilault2020conditionally, pfeiffer2020adapterfusion, wang2020kadapter, wu2020understanding}.

To address these limitations, we propose a novel, modularized framework that aims to learn multiple expert models for KGs, then conduct zero-shot fusion to allow collaboration among KGs. For this purpose, we leverage AdapterFusion~\citep{pfeiffer2020adapterfusion} where multiple tiny modules between Transformer blocks called adapters~\citep{houlsby2019parameter} can be combined after independent training, thus allowing a continual integration of the adapters without retraining the entire framework. Specifically, we treat the adapters as different KG-specific experts, and combine them using an attention-like fusion module. To improve the fusion of adapters, we suggest a KG-alignment adapter that guides to the apt \emph{expert adapters}. Here, we use KGs in three different synthetic supervision training: (1) KG-specific QA datasets to train the KG-specific expert adapters, (2) a KG classification datasets to train the KG-alignment adapter, and (3) a balanced mixture of KG-specific QA datasets to train the fusion module. Our modularized method alleviates the interference between different KGs, which is the pitfall of MTL from our empirical observation, and thus combines multiple KGs into a synergetic zero-shot framework.

Our contributions are: (1) We suggest a simple, yet effective KG modularization strategy for the use of multiple KGs in commonsense reasoning. (2) We then explore the use of AdapterFusion~\citep{pfeiffer2020adapterfusion} for better knowledge aggregation based on the KG modularization in zero-shot setting. We believe that such modularized transfer learning is critical to using different knowledge sources synergetically against interference between them. (3) In extensive experiments on various commonsense reasoning benchmarks, our framework achieves significant improvements over baselines using a single KG, even using multiple KGs, which implies the robustness in commonsense reasoning.

\section{Related Work \& Preliminaries}

\subsection{Zero-shot Commonsense Reasoning}

Many researchers have recently focused on building unsupervised models without any benchmark supervisions (\ie, zero-shot learning). In such zero-shot setting, KGs are often used as an external resource for improving model prior (\eg, continually learned from pre-trained language models)~\citep{banerjee2020self,bosselut2019dynamic,ma2020knowledgedriven}, especially for commonsense reasoning, as much existing work couples language models with neural/symbolic commonsense KGs.

However, most of existing work are either assuming the existence of the alignment information between tasks and KGs~\citep{banerjee2020self} or an integrated KG~\citep{ma2020knowledgedriven}. For example, $\texttt{ATOMIC}^{20}_{20}$~\citep{hwang2020comet}, a commonsense KG which incorporates tuples from \texttt{ConceptNet} and \texttt{ATOMIC} with new relations and further crowdsourcing, combines multiple KGs into a new integrated KG, but as widely known~\citep{ilievski2020consolidating, hwang2020comet}, heterogeneous schema between different KGs may limit triplets that can be integrated.\footnote{Only 172K tuples of the 3.4M tuples and 5 relations of 36 relations in \texttt{ConceptNet} are integrated into $\texttt{ATOMIC}^{20}_{20}$.} Rather than such symbolic KG integration with the inevitable loss of knowledge, in this work, we explore the neural KG integration leveraging the multiple KGs without additional processing and alignment information between KG and task.

\subsection{Transfer Learning with Modular Approaches}
The idea of having specialized parameters, or so-called experts, has been widely studied to integrate multiple sources of knowledge via transfer learning. The adapter module~\citep{rebuffi2017learning,houlsby2019parameter} has been explored as one of such approaches, introducing a small number of task-specific parameters at every layer of pre-trained language model (PLM) while sharing the parameters of underlying PLM which is fixed. To address the limitations of transfer learning due to high re-training cost, many works utilize the multiple adapter modules for individual tasks with different domains~\citep{puigcerver2020scalable, bapna2019simple, ruckle2020multicqa, madotto2020adapterbot} considering each adapter to be an expert of each domain. Similar to our work, K-Adapter~\citep{wang2020kadapter} encodes factual and linguistic knowledge to each adapter, but in this paper, we further explore how to mitigate catastrophic forgetting or interference among multiple adapters for better knowledge transfer in zero-shot setting.

\subsection{Multi-Task Learning}

MTL~\citep{liu2019multi,zhang2017surveymtl,caruana1997multitask} learns a shared representation while aggregating knowledge across multiple learning tasks, often leading to better generalization ability of a model. However, parametric aggregation of knowledge with MTL has following limitations: (1) retraining the full model when adding new tasks~\citep{houlsby2019parameter,pfeiffer2020adapterfusion,pfeiffer2020low} (2) catastrophic forgetting and interference between tasks leading to difficulties of solving each task equally well~\citep{pilault2020conditionally,wu2020understanding,yu2020gradient} and (3) inconsistent effect~\citep{lourie2021unicorn}. To deal with these challenges,  Mixture-of-Experts (MoE) is a parameterized generalization of ensembling techniques, which has been adapted for MTL with gating network trained to optimize each task~\citep{ma2018modeling}. However, simple linear gating networks are too shallow and thus may destruct task knowledge for commonsense reasoning. 

To address this problem, AdapterFusion~\citep{pfeiffer2020adapterfusion} has been proposed to fuse task specific parameters called adapters for the given target task leveraging attention-like mechanism. AdapterFusion aggregates adapters, which is trained independently for each task, in a non-destructive manner mitigating aforementioned MTL problems such as forgetting and interference between tasks. Recently, it has been used for zero-shot cross-lingual transfer framework~\citep{pfeiffer2020mad, wang2021efficient}, which motivates our work to transfer multi-source knowledge with less interference for zero-shot commonsense reasoning.

\section{Modularized Zero-shot Framework}\label{sec:framework}
In our setup, we repurpose synthetic QA generation~\citep{ma2020knowledgedriven} for the task of knowledge-driven zero-shot learning for commonsense reasoning, \ie, we transform a KG into multiple $(Q_i, A_i)$ pairs where $Q_i$ is a natural language question and $A_i=\{A_{i,1}, ..., A_{i,m}\}$ is the set of options with $m$ answer candidates. Specifically, given a triple $(e^{head}, r, e^{tail})$ in a KG, where $e^{head}$, $e^{tail}$ and $r$ denote head/tail entity and relation respectively, we transform $e^{head}$ and $r$ into a natural language question $Q_i$ using templates. For the option set $A_i$, we use the combination of the correct answer $e^{tail}$ and $m-1$ distractors which are tail entities from other triples sampled randomly~\citep{ma2020knowledgedriven}. Details are described in Appendix~\ref{ap:synthetic_qa}. 

\begin{table}[t!] \small
\renewcommand{\arraystretch}{1.0}
\begin{tabular}{l} 
\hline
\multicolumn{1}{p{0.45\textwidth}}{\raggedright \textbf{QA from} \texttt{ATOMIC}~\citep{sap2019atomic}} \\ \hline
\multicolumn{1}{p{0.45\textwidth}}{\raggedright Q: Dana speeds on the highway. Dana is seen as} \\ 
\multicolumn{1}{p{0.45\textwidth}}{\raggedright A1: considerate \textbf{A2: risky}(\checkmark) A3: lazy} \\ 
\hline

\multicolumn{1}{p{0.45\textwidth}}{\raggedright \textbf{QA from} \texttt{ConceptNet}~\citep{speer2017conceptnet}} \\ \hline
\multicolumn{1}{p{0.45\textwidth}}{\raggedright Q: pentode is a [MASK]} \\ 
\multicolumn{1}{p{0.45\textwidth}}{\raggedright A1: ascocarp A2: girls footwear \textbf{A3: vacuum tube}(\checkmark)} \\ 
\hline

\multicolumn{1}{p{0.45\textwidth}}{\raggedright \textbf{QA from} \texttt{WikiData}~\citep{wikidata}} \\ \hline
\multicolumn{1}{p{0.45\textwidth}}{\raggedright Q: badminton is a [MASK]} \\ 
\multicolumn{1}{p{0.45\textwidth}}{\raggedright A1: fable  A2: juvenile justice \textbf{A3: type of sport}(\checkmark)} \\ 
\hline

\multicolumn{1}{p{0.45\textwidth}}{\raggedright \textbf{QA from} \texttt{WordNet}~\citep{miller1995wordnet}} \\ \hline
\multicolumn{1}{p{0.45\textwidth}}{\raggedright Q: princewood is part of [MASK]} \\ 
\multicolumn{1}{p{0.45\textwidth}}{\raggedright A1: shaddock \textbf{A2: genus Cordia}(\checkmark) \\ A3: family Columbidae} \\ 
\hline
\end{tabular}

\caption{Synthetic QA examples. We use templates to convert ($e^{head},r$) into a natural language sentence.} 
\label{table:1-syntheticqa-sample}
\end{table}

Formally, we denote $(Q_i,A_i)$ as one QA sample. The goal is to learn a QA model from the synthetic QA sample. In a downstream task (\eg, reasoning benchmarks such as SocialIQA and CommonsenseQA), we need to predict answers given non-synthetic test samples $(Q^{test},A^{test})$. In the training stage, we are given $K$ KG-driven datasets $\{\calD^k_{QA}\}_{k=1}^K$ from $K$ different KGs, where $\calD^k_{QA}$ is a dataset with $N_k$ samples for KG $k$ as follows:
\begin{equation}\label{eq:qadata}
\calD^k_{QA}  = \{(Q_i,A_i,label)\}_{i=1}^{N_k}
\end{equation}
where $label$ is the index of the correct answer for each sample. In this work, as shown in Table~\ref{table:1-syntheticqa-sample}, we generate four synthetic QA datasets from \texttt{ATOMIC}, \texttt{ConceptNet}, \texttt{WikiData}, and \texttt{WordNet} (More details are in Appendix~\ref{ap:commonsense_knowledge_graph}).

\begin{figure}[t!]
\centering
    \includegraphics[width=\columnwidth]{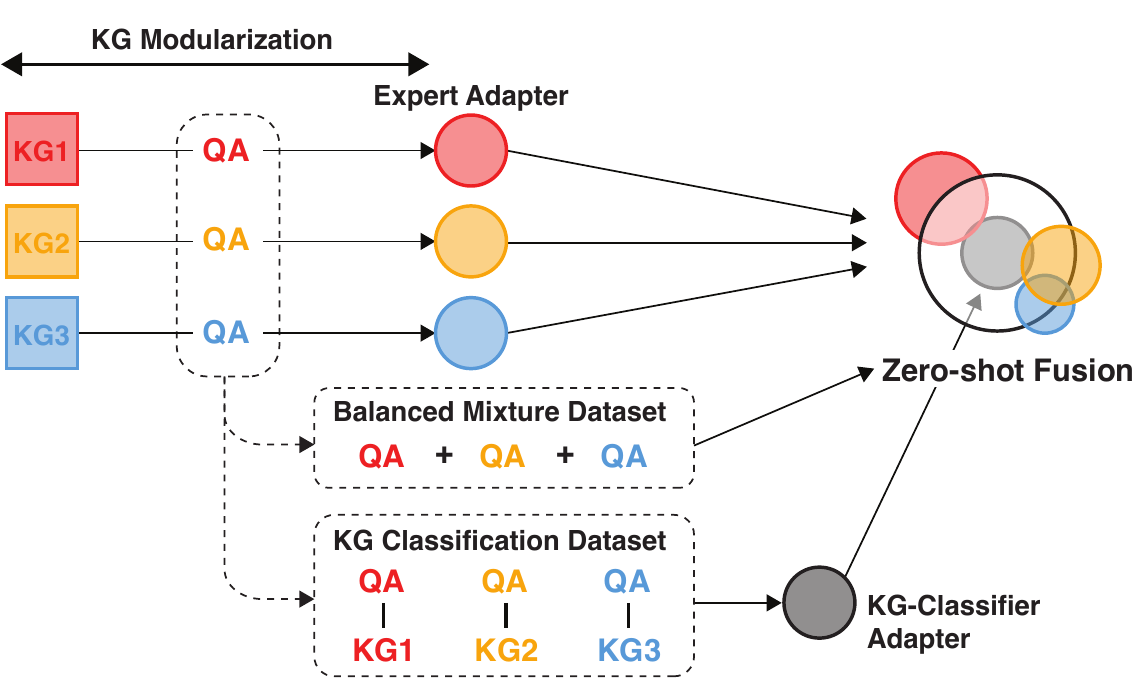}
\caption{Illustration of the proposed modularized framework for zero-shot commonsense reasoning. Each colored square represents different KGs. Not only for KG modularization, we re-use a set of synthetic QA datasets from the multiple KGs for the purpose of KG classification and zero-shot fusion, which enables better knowledge aggregation.}
\label{fig:framework-illustration}
\end{figure}

For effective use of multiple KGs at once with less interference, 
we present a modularized framework, which is a novel approach to knowledge transfer for the zero-shot setting as shown in Figure~\ref{fig:framework-illustration}. As a modular approach, we train the multiple KG-specific adapters (\emph{expert adapters}) with each dataset from KG. Based on these pre-trained adapters, we use a zero-shot fusion method to aggregate knowledge of each adapter leveraging AdapterFusion~\citep{pfeiffer2020adapterfusion} as a base architecture with the balanced mixture of each KG dataset. Further, for better knowledge fusion, we suggest a KG-alignment aware adapter (\emph{KG-Classifier adapter}) as a guide for detecting alignment with given sample in zero-shot reasoning. Here, we utilize KG classification dataset by verifying the synthetic QAs. Algorithm~\ref{alg:ourframework} in Appendix outlines the overall process of our proposed framework. We summarize the notations in Appendix~\ref{ap:notation}.

\subsection{KG Modularization}\label{sec:framework.1}
First, we modularize the KGs to preserve their intrinsic knowledge. Considering the importance of using a suitable and well-aligned KG~\citep{ma2019towards, ma2020knowledgedriven} on a downstream task, the subtle difference between each KG should be learned by the model without any interference from each other. Accordingly, we adopt the adapter module~\citep{houlsby2019parameter} which repurposes a pre-trained language model (PLM) to incorporate each KG as tiny modules in between Transformer blocks. Specifically, as illustrated in Figure~\ref{fig:module-figure} (except for green area), the adapter training strategy involves injecting new layers (parameterized by $\Phi$) into the original PLM (parameterized by $\theta$). The weights of the original PLM are untouched, while the new adapter layers are initialized at random. Formally, we call each adapter trained with $\calD^k_{QA}$ as an \emph{expert adapter} for KG $k$, parameterized by $\Phi_{QA}^k$. 


When a QA sample $(Q_i,A_i)$ is given for dataset $\calD_{QA}^k$, we first concatenate question $Q_i$ and each answer option $A_i=\{A_{i,1},...,A_{i,m}\}$ to generate input sequences $T_i=\{T_{i,1},...,T_{i,m}\}$. Then, we compute a score $S_{i,j}$~\citep{ma2020knowledgedriven} for the answer candidate $A_{i,j}$ is computed as follows:
\begin{equation}\label{eqn:score_qa}
    S_{i,j} = -\frac{1}{|T_{i,j}|}\sum_{t=1}^{|T_{i,j}|}logP(w_t | ...w_{t-1},w_{t+1}...; \theta, \Phi)
\end{equation}
where $w_t$ is a word token in the sequence $T_{i,j}$ and $P$ is the conditional probability from Transformer blocks parameterized by $\theta$ and $\Phi$. To train the adapter $\Phi_{QA}^k$, we use the marginal ranking loss~\citep{ma2020knowledgedriven} as follows:
\begin{equation}
    \calL_{QA} = \frac{1}{m} \sum_{i=1}^{N_k} \sum_{\substack{j=1 \\ j \neq label}}^{m}max(0, \eta - S_{i, label} + S_{i, j})
\end{equation}
where $\eta$ represents the margin. 
\begin{equation}\label{eq:qamodel}
    \Phi_{QA}^k \leftarrow \operatorname*{argmin}_{\Phi} 
    \calL_{QA}(\mathcal{D}^{k}_{QA} ; \theta, \Phi)
\end{equation}
where KG-invariant parameters $\theta$ are fixed and only KG-dependent parameters $\Phi_{QA}^k$ are learned, which enables to store the corresponding knowledge separately without any interference. Further, we can parallelize the training of adapter for all KGs. The efficiency of adapter training allows our modularization to be more scalable.

\subsection{Zero-shot Fusion}

\begin{figure}[t]
\centering
    \includegraphics[width=\columnwidth]{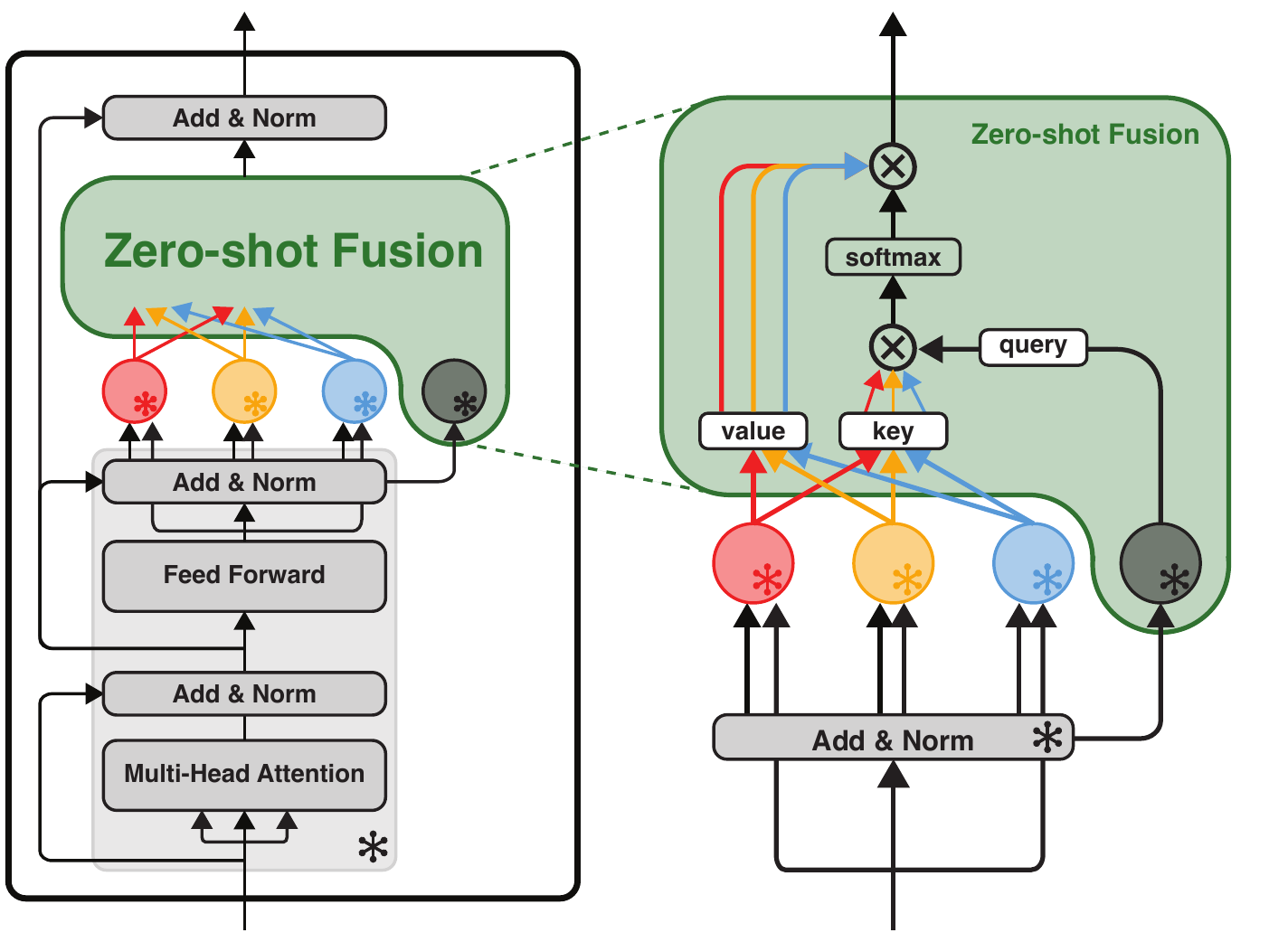}
\caption{Illustration of the zero-shot fusion architecture with \emph{KG-Classifier adapter}. Each colored circle represents \emph{expert adapters}, except the black circle which denotes \emph{KG-Classifier adapter}. $\ast$ indicates the fixed layer. Details are in Appendix~\ref{ap:module_figure_parameter}}
\label{fig:module-figure}
\end{figure}

Once the expert adapters are learned, we combine the knowledge from each expert adapter using an attention-like mechanism. We present a novel fusion strategy as shown in Figure~\ref{fig:module-figure}, which is referred to as the zero-shot fusion. In contrast to AdapterFusion~\citep{pfeiffer2020adapterfusion} 
where the focus is learning to transfer knowledge to a specific target task, our zero-shot fusion aims to generalize this transfer to any arbitrary target task.
Specifically, the zero-shot fusion parameters $\Psi$ learn to combine fixed expert adapters which are parameterized by $\Phi_{QA}^1,...,\Phi_{QA}^K$.
In each Transformer layer $l$ of PLM with the injected fusion layer, the zero-shot fusion parameters $\Psi_{QA}$ consist of query, key, and value matrices, denoted by $\textbf{W}_l^Q$, $\textbf{W}_l^{K}$, and $\textbf{W}_l^{V}$ respectively. These parameters are used to learn the balancing between the representation of each \emph{expert adapters} through attention-like mechanism. While fixing both the parameters $\theta$ and all expert adapters $\Phi_{QA}^1,...,\Phi_{QA}^K$, the only trainable weights $\Psi_{QA}$ on the fusion layer learns to combine the knowledge from different $K$ \emph{expert adapters} by using the subset of $\{\calD_{QA}^k\}_{k=1}^K$ by random sampling. Here, we balance the ratio between the $K$ knowledge-driven datasets as $N$ samples (details are in Appendix~\ref{ap:dataset_for_zeroshot_fusion}). Formally,
\begin{equation}\label{eq:fusionmodel}
    \Psi_{QA} \leftarrow \operatorname*{argmin}_{\Psi} \sum_{k=1}^{K} \calL_{QA}(\mathcal{D}^k_{QA}; \theta, \{ \Phi_{QA}^k\}_{k=1}^K, \Psi)
\end{equation}
where $\Psi$ refers to the initialized zero-shot fusion parameters.

More specifically, in the $l$-th Transformer layer, let $h_{PLM}^{l}$ and $h_{E}^{k,l}$ be the representations of underlying PLM parameterized by $\theta$ and an \emph{expert adapter} parameterized by $\Phi_{QA}^k$, respectively. Then, using the hidden representation $h_{PLM}^{l}$ of PLM as a query, the fusion layer performs the attention-like function as follows:
\begin{align}
    \textbf{K}_l,\textbf{V}_l &= [h_{E}^{1,l},...,h_{E}^{K,l}] \\
    \textbf{Q}_l &= h_{PLM}^l\label{eq:5} \\
    \textbf{z}_l &= \text{Attention}(\textbf{Q}_l\textbf{W}_l^Q,\textbf{K}_l\textbf{W}_l^K,\textbf{V}_l\textbf{W}_l^V)
\end{align}
where $\textbf{z}_l$ is passed to the next Transformer layer. Given a sample, the zero-shot fusion learns the suitable balancing parameters between the \emph{expert adapters} for zero-shot reasoning. Eventually, it learns to identify generalizability across commonsense reasoning tasks. 

\subsection{KG-Classifier Adapter}\label{sec:framework.2}

AdapterFusion uses the PLM hidden representation $h_{PLM}^l$ as a query which is learned when training on a specific downstream task. In our zero-shot setting, however, we use a mixture of synthetic QA for fusion training, which is not exactly a training dataset for a downstream task.
To compensate for this issue, we present \emph{KG-Classifier adapter}, which is a KG alignment-aware adapter, which is motivated from the fact that the ability to find which KG has an alignment with the given sample can be helpful as a role of providing a guidance for better performance~\citep{ma2019towards, ma2020knowledgedriven}.

Specifically, we propose a novel training task for \emph{KG-Classifier adapter}, which requires predicting the KG for the given sample of the task. For that, given $\{\calD_{QA}^k\}_{k=1}^K$, we first transform a QA sample $(Q_i,A_i)$ into a new KG classification sample $[Q_i;A_{i,label}]$ where $[;]$ is the concatenation. Then, we obtain a new label $y_i \in \{0,1\}^K$ indicating the corresponding KG source. The samples are in Appendix~\ref{ap:KG_classification_dataset}. Formally, KG classification dataset $\calD_{KGC}$ is defined as:
\begin{equation}\label{eq:kgcdata}
    \calD_{KGC} = \{([Q_i;A_{i,label}] , y_i)\}_{i=1}^{M}
\end{equation}
where $M$ is the total size of $\{\calD_{QA}^k\}_{k=1}^K$.


Based on $\calD_{KGC}$, we learn the \emph{KG-Classifier adapter} parameterized by $\theta$ and $\Phi_{KGC}$. First, a classification sample $i$ is encoded into $h_{CLS} \in \mathbb{R}^{H}$ then scored as $\hat{y}_{i} \in \mathbb{R}^K$ with a linear layer $W_{KGC} \in \mathbb{R}^{K \times H}$, \ie, $\hat{y}_{i} = W_{KGC} h_{CLS}$. Once $\hat{y}_i$ is normalized by a softmax layer, the network is trained to minimize the cross-entropy loss $\calL_{KGC}$ between the prediction $\hat{y}_i$ and its ground truth $y_i$:
\begin{equation}\label{eq:kgcmodel}
    \Phi_{KGC} \leftarrow \operatorname*{argmin}_{\Phi} \sum_{i=1}^{M} \calL_{KGC}(y_i, \hat{y}_i ; \theta, \Phi)
\end{equation}

We propose to use the representation of \emph{KG-Classifier adapter} as a query in attention-like mechanism, referred to as the zero-shot fusion with \emph{KG-Classifier adapter}. That is, using the hidden representation $h_{KGC}^{l}$ of a \emph{KG-Classifier adapter} parameterized by $\Phi_{KGC}$ as a query, we substitute $\textbf{Q}_l$ in Eq. (\ref{eqn:mean_query}) as follows: 
\begin{equation}\label{eqn:mean_query}
    \textbf{Q}_l = h_{KGC}^l
\end{equation}
The overall zero-shot fusion architecture including \emph{KG-Classifier} is illustrated in Figure~\ref{fig:module-figure}.

\begin{table*}[t!]\small
\renewcommand{\arraystretch}{1.15}
\centering
\scalebox{0.83}{
\begin{tabular}{l c c c c c c c}
 \hline
 \noalign{\hrule height 0.8pt}
    \textbf{Model} & \textbf{KG} & \textbf{a-NLI}  & \textbf{CSQA} & \textbf{PIQA} & \textbf{SIQA} & \textbf{WG} & \textbf{Avg.}\\
\hline
\noalign{\hrule height0.8pt}
    Random      &   -    & 50.0 & 20.0 & 50.0 & 33.3 & 50.0 & 40.7\\
    Majority      &   -    & 50.8 & 20.9 & 50.5 & 33.6 & 50.4 & 41.2\\ 
    GPT2-L        &   -    & 56.5 & 41.4 & 68.9 & 44.6 & 53.2 & 52.9\\ 
    RoBERTa-L     &   -    & 65.5 & 45.0 & 67.6 & 47.3 & 57.5 & 56.6 \\ 
    Self-talk \citep{shwartz2020unsupervisedselftalk}    &   -    &   -  & 32.4 & 70.2 & 46.2 & 54.7 & - \\
    COMET-DynaGen ~\citep{bosselut2019dynamic} & AT & - & - & - & 50.1 & - & - \\
    SMLM ~\citep{banerjee2020self}          &   *    & 65.3 & 38.8 &   -  & 48.5 &   -  & -\\
    RoBERTa-L (MR) \citep{ma2020knowledgedriven} &   AT    & 70.8 & 64.2 &  72.1 & 63.1 & 59.2 & 65.9 \\
    RoBERTa-L (MR) \citep{ma2020knowledgedriven} &   CN,WD,WN    & 70.0 & 67.9 & 72.0 & 54.8 & 59.4 & 64.8 \\
    RoBERTa-L (MR) \citep{ma2020knowledgedriven} &   Whole    & 70.5 & 67.4 & 72.4 & 63.2 & \textbf{60.9} & 66.9 \\
\hline
\noalign{\hrule height0.8pt}
    MTL         & Whole & 69.8 (± 0.5) & 66.0 (± 0.9) &	71.2 (± 0.8) &	62.2 (± 1.0) &	59.5 (± 0.2) &	65.7 \\
\noalign{\hrule height0.8pt}
    \textbf{zero-shot fusion w/o \emph{KG-C adapter}}     & Whole & 72.3($\pm$0.4) & 67.9($\pm$0.2) & \textbf{73.1}($\pm$0.4) & 65.9($\pm$0.5) & 59.7($\pm$0.2) & 67.8 \\
    \textbf{zero-shot fusion w/ \emph{KG-C adapter}}     & Whole & \textbf{72.5}($\pm$0.2) & \textbf{68.2}($\pm$0.2) & 72.9($\pm$0.4) & \textbf{66.6}($\pm$0.1) & 60.8($\pm$0.1) & \textbf{68.2}\\
    
  \hline
 \noalign{\hrule height 0.8pt}
 \end{tabular}}
 
\caption{Zero-shot evaluation results with different combinations of models and knowledge sources across five commonsense tasks. AT, CN, WD and WN represent \texttt{ATOMIC}, \texttt{ConceptNet}, \texttt{WikiData} and \texttt{WordNet}, respectively. Whole represents the combination of AT, CN, WD and WN. Bold text indicates the best performance on each benchmark. RoBERTa-L (MR) used the synthetic dataset after filtering, while we use the raw version. SMLM (*) used different KG which has strong alignment with each task (\eg AT for SIQA).
} 
\label{table:main_result}
\end{table*}

\section{Experiments}
In this section we evaluate the efficacy of our framework on five commonsense reasoning tasks. We denote \emph{KG-Classifier adapter} by \emph{KG-C adapter}.

\subsection{Experimental Settings} 
All our experiments are conducted in a zero-shot setting, in which the models do not have access to the official training data or labels of the benchmark. For the evaluation, we use the validation set of each benchmark\footnote{Since the official test sets are not publicly available}, however, the validation set of each benchmark can be role as an test set since it is not used for hyperparameter tuning or model selection. We use accuracy as a metric.

\subsubsection{Benchmarks} We evaluate our proposed framework on five question-answering benchmarks for commonsense reasoning: SocialIQA (SIQA)~\citep{sap2019socialiqa}, CommonsenseQA (CSQA)~\citep{talmor2018commonsenseqa}, Abductive NLI (a-NLI)~\citep{bhagavatula2019abductive}, PhysicalIQA (PIQA)~\citep{bisk2020piqa}, and WinoGrande (WG)~\citep{sakaguchi2020winogrande}. Each commonsense benchmark evaluates a specific kind of knowledge: social commonsense for SIQA, concept-level commonsense for CSQA, abductive reasoning for a-NLI, physical commonsense for PIQA, and pronoun resolution ability for WG.\footnote{Some benchmarks have a strong alignment with a certain KG due to its construction strategy: SIQA-\texttt{ATOMIC}, and CSQA-\texttt{ConceptNet}. To make a direct comparison with~\citet{ma2020knowledgedriven}, we use the same KGs to generate data samples.} The details are presented in Appendix~\ref{ap:commonsense_benchmarks}. 
 
\subsubsection{Baselines} 
We compare our framework with the following baselines. First, to show the characteristics of each benchmark, we use the random or the most frequent label as \emph{Random} and \emph{Majority} baseline, respectively. RoBERTa-L and GPT2-L is the performance of each PLM without any fine-tuning. Also, as the baseline for the unsupervised learning model using KGs, we report the performance of Self-talk~\citep{shwartz2020unsupervisedselftalk}, COMET-DynaGen~\citep{bosselut2019dynamic}, SMLM~\citep{banerjee2020self} as presented in original papers. 

For further analysis in $\S$\ref{sec:mitigating_interference} and $\S$\ref{sec:visualization_knowledge_Aggregation}, we set the following models that are pre-trained on the synthetic QA datasets from KGs as baselines:
\begin{itemize}
    \item \textbf{Single-Task Learning (STL)}: The model is pre-trained on a synthetic QA dataset generated from a single KG. Specifically, we experiment two architectural choices: PLM (STL-PLM) and PLM with adapters (STL-Adapter). For each architecture, there are four STL models for each of synthetic QA datasets derived from \texttt{ATOMIC}, \texttt{ConceptNet}, \texttt{WikiData}, and \texttt{WordNet}. We note that the trained STL-Adapter is an \emph{expert adapter} from a specific KG in our framework. The performance of each STL baseline is shown in Appendix~\ref{ap:zeroshot_fusion_wo_kgc_relative_performance} Table~\ref{table:mtl_result} and Table~\ref{table:zeroshot_fusion_kgc_result}.
    \item \textbf{Multi-Task Learning (MTL)}: The model is pre-trained on multiple synthetic QA datasets, each of which is generated from a KG. We experiment with a PLM trained on all four aforementioned synthetic QA datasets. We note that the difference between STL-PLM and MTL is whether to use one synthetic QA dataset or multiple synthetic QA datasets for its training.
\end{itemize}

\subsubsection{Implementations} We employ RoBERTa-L~\citep{liu2019roberta} from Hugging Face’s transformers toolkit for all experiments. We follow the default settings from ~\citet{ma2020knowledgedriven}. Our implementation uses Adapter~\citep{houlsby2019parameter} and AdapterFusion~\citep{pfeiffer2020adapterfusion} as a base model architecture from AdpaterHub~\citep{pfeiffer2020adapterhub}. We run our experiments with three different random seeds. The implementation details are described in Appendix~\ref{ap:implementation_details}.

\subsection{Main Results} 

Table~\ref{table:main_result} shows the zero-shot evaluation results on five benchmark datasets. Generally, zero-shot fusion scores higher than the baselines across all benchmarks, and further, zero-shot fusion shows the best performance in all benchmarks except WG. We note that although \citet{ma2020knowledgedriven} uses the synthetic QA dataset after sample filtering, our method achieves comparable performance with the best performance in WG, even with the raw dataset. Also, the average score of all evaluation benchmarks (the last column of Table~\ref{table:main_result}) shows that zero-shot fusion has generalisability in commonsense reasoning. 

In addition, zero-shot fusion achieves consistent improvements over MTL. These results indicate that our proposed zero-shot fusion method attributes to fusing the knowledge of multiple KGs more synergetically regardless of the task.

Moreover, as an ablation, we compare the zero-shot fusion with and without \emph{KG-C adapter} to explore the efficacy of the \emph{KG-C adapter}. We can observe that zero-shot fusion with \emph{KG-C adapter} improves the average accuracy by 0.4\%, which implies that the use of \emph{KG-C adapter} improves the overall performance and makes our method generalize better on most of the evaluation benchmarks.

\subsection{Impact of the KG-Classifier Adapter}\label{subsec:KG-classifier-impact}

\begin{figure}[t!]\small
    \begin{center}
    \begin{tabular}{@{} c c@{}}
    \includegraphics[width=0.45\columnwidth]{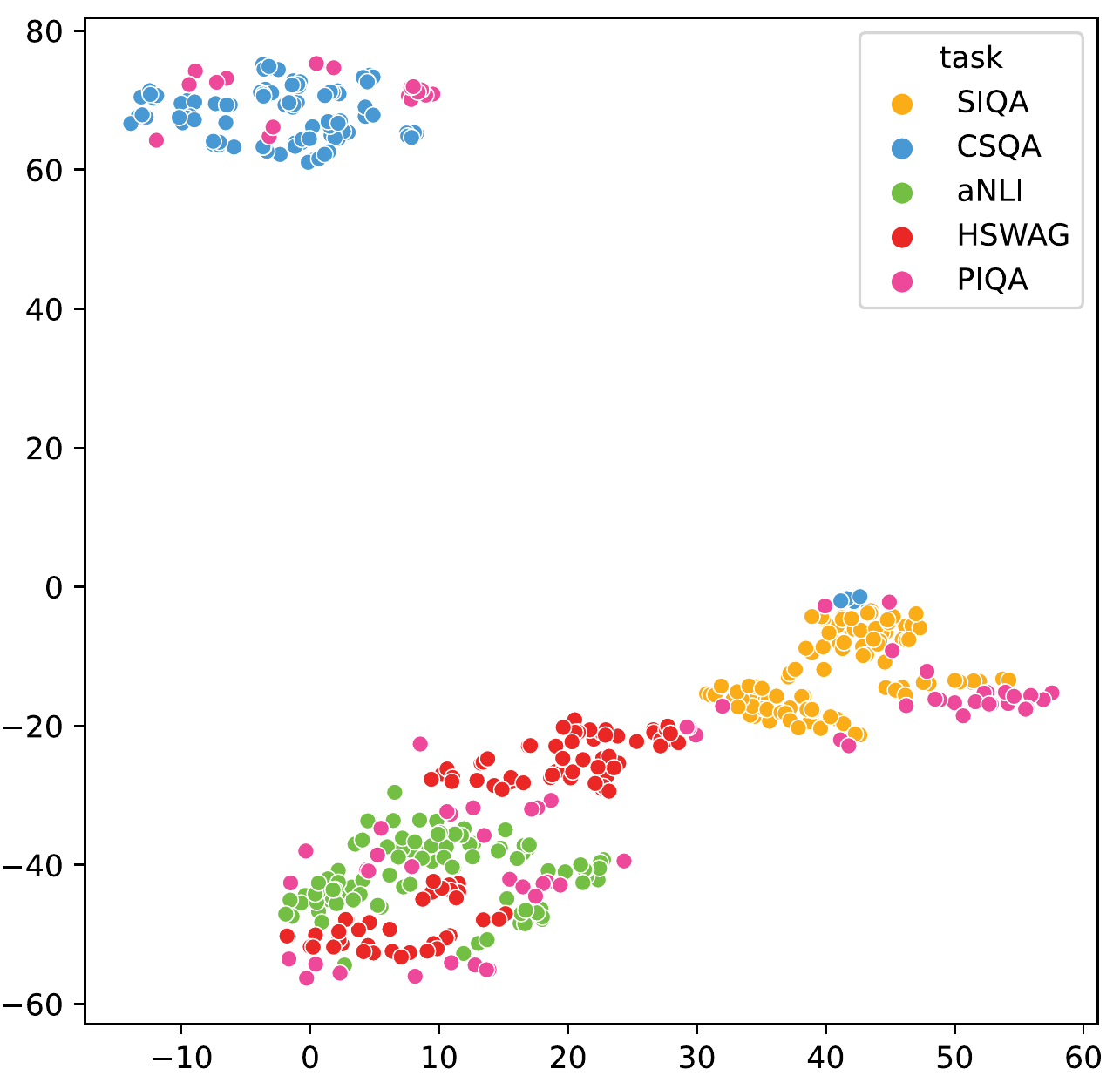} & 
    \includegraphics[width=0.45\columnwidth]{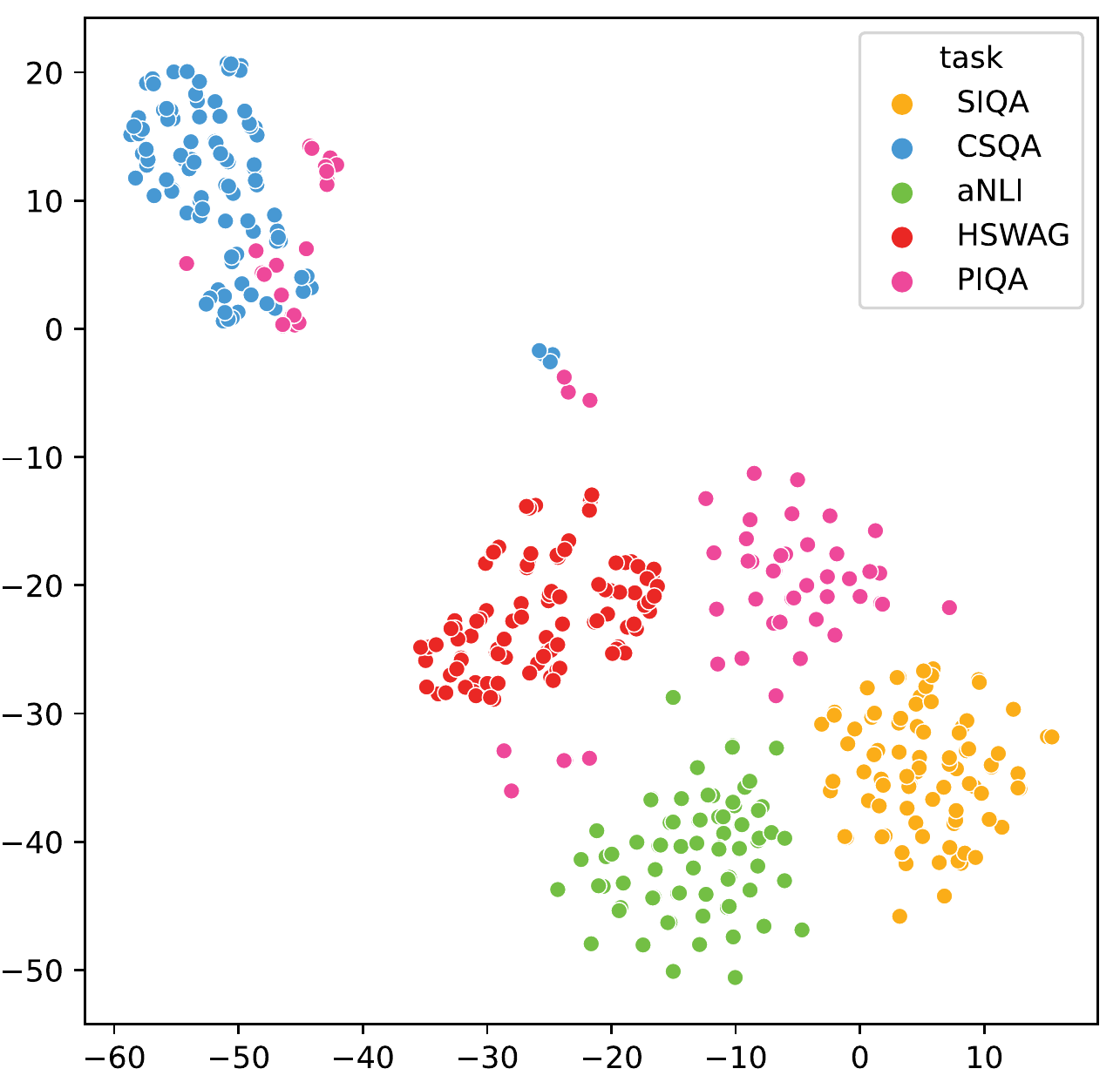} \\
    (a) PLM & (b) \emph{KG-Classifier adapter} \\
    \end{tabular}
    \end{center}
    \caption{t-SNE visualization of the hidden representation from (a) PLM and (b) \emph{KG-C adapter}. Each color denotes the five different benchmark samples.}
    \label{fig:kg-classifier}
\end{figure}

To assess the effects of the \emph{KG-C adapter} itself, we visualize and compare the final layer \texttt{[CLS]} token representation between PLM and \emph{KG-C adapter}. 
Figure~\ref{fig:kg-classifier} shows t-SNE~\citep{van2008visualizingtsne} plots of all representation of five benchmark datasets. In this figure, every sample is mapped into a 1024-dimensional feature space through RoBERTa-L model and projected back into a two-dimensional plane by t-SNE. We can observe that \emph{KG-C adapter} can separate the samples of different benchmarks well despite being unseen data. It verifies that KG-awareness acquired with the KG classification task is beneficial to categorize the given sample. The \emph{KG-C adapter} can thus generate a relevant KG-aware query for a given sample and help to fuse representations from suitable \emph{expert adapters} in our proposed framework.

\begin{figure}[t!]\small
    \begin{center}
    \begin{tabular}{@{}c c@{}}
    \includegraphics[width=0.48\columnwidth]{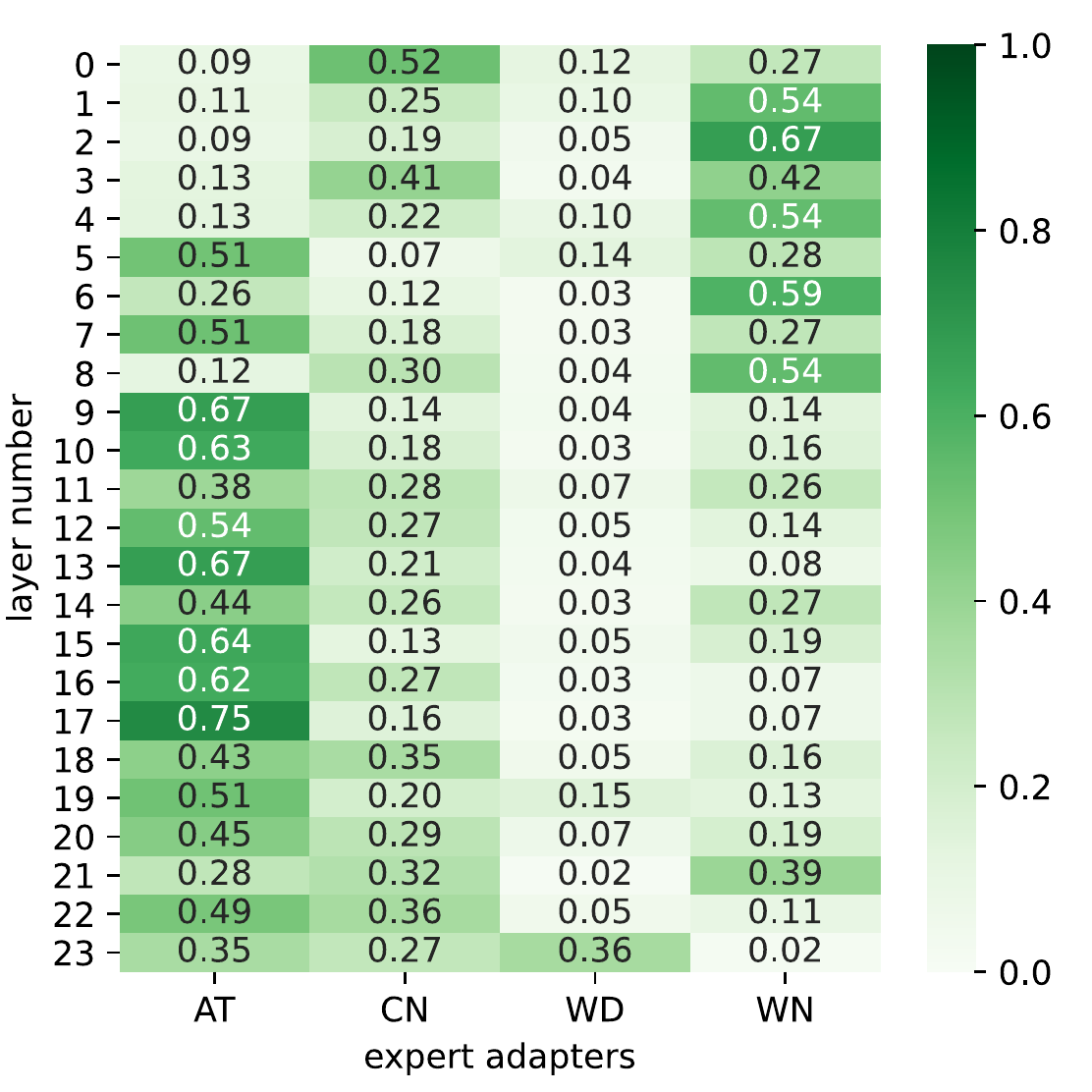} & 
    \includegraphics[width=0.48\columnwidth]{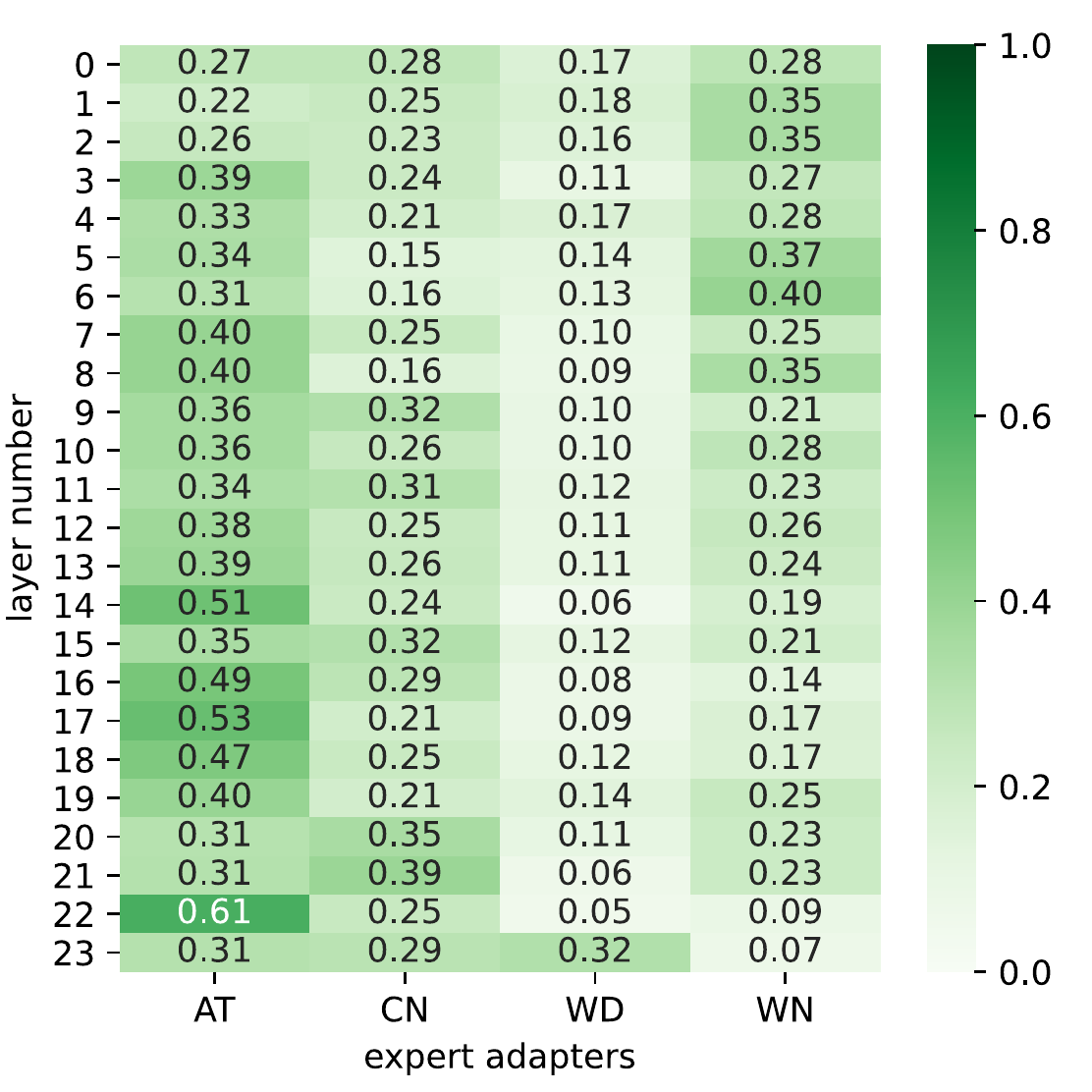} \\
    (a) w/o \emph{KG-C adapter}  & (b) w/\emph{KG-C adapter}  \\
    \end{tabular}
    \end{center}
    \caption{
    Comparison of attention probability between zero-shot fusion with/without \emph{KG-C adapter}. The x- and y-axis indicate \emph{expert adapters} and the fusion layer number in RoBERTa-L, respectively. The darker color indicates higher attention probability in fusion layer.
    }
    \label{fig:attention-probs}
\end{figure}

Further, we explore how the \emph{KG-C adapter} affects zero-shot fusion which is based on an attention-like mechanism~\citep{pfeiffer2020adapterfusion} compared to zero-shot fusion without \emph{KG-C adapter}. Here, while zero-shot fusion without \emph{KG-C adapter} simply uses the representation of PLM as a query, zero-shot fusion with \emph{KG-C adapter} leverages the representation of \emph{KG-C adapter}. To illustrate this strength, we visualize the attention probability of \texttt{[CLS]} token from each fusion layer as a representative in Figure~\ref{fig:attention-probs}. The column of the darker cell indicates the adapter that has the bigger influence on the fused representation. We can observe that zero-shot fusion with \emph{KG-C adapter} fuses the knowledge from different experts with a subtle difference rather than focusing on a single expert severely. This implies that \emph{KG-C adapter} enables the delicate balancing between multiple knowledge sources based on the KG-alignment awareness, which leads to performance improvements in commonsense reasoning tasks. Interestingly, both cases have the ability not to focus on the \emph{expert adapter} based on \texttt{WikiData}, which can be seen as a redundant expert.\footnote{The zero-shot fusion with \emph{KG-C adapter} using AT, CN, and WN shows the best average performance in Table~\ref{table:zeroshot_fusion_kgc_result}.} This observation would benefit from the further study that explores the optimal combination of KGs by expert selection or rejection.

\subsection{Mitigating Interference}\label{sec:mitigating_interference}
In this experiment, we compare the amount of interference in the MTL and zero-shot fusion with \emph{KG-C adapter}. We propose a novel evaluation metric, the \textit{interference ratio}, which is the percentage of the incorrectly predicted samples by the multi-KG models among the correctly predicted samples from the STL models in common.

\begin{figure}[t]
\centering
    \includegraphics[width=\columnwidth]{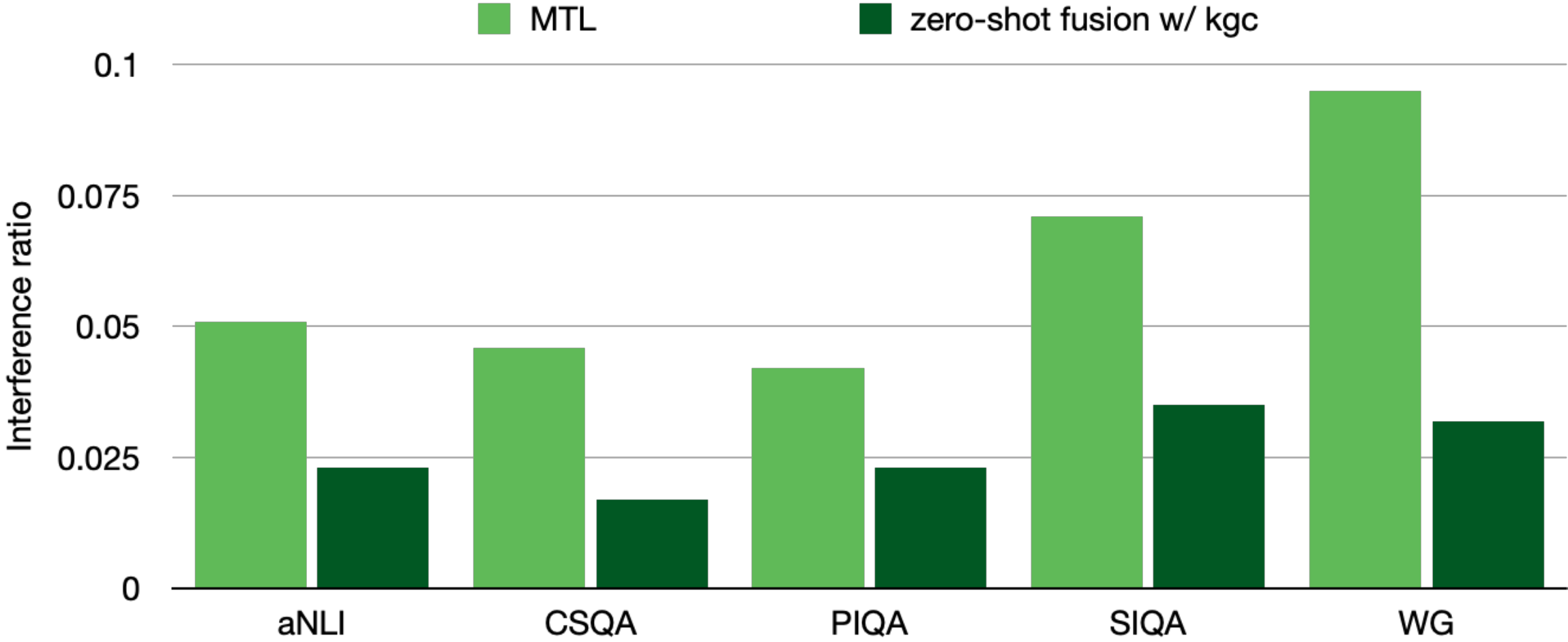}
\caption{Interference ratio of multi-KG models on five benchmarks. The lower indicates less interference.}
\label{fig:interference-ratio-all}
\end{figure}

Using the interference ratio, we can precisely compare the negative effects of multi-KG models on knowledge aggregation since the only reason to get the correct samples wrong is the interference caused by learning with additional KGs. We present the interference ratio of the models on five benchmark datasets in Figure~\ref{fig:interference-ratio-all}. This figure shows that MTL has the higher interference ratio than the competing models across all benchmarks. Our method achieves a substantially better ratio, especially when \emph{KG-C adapter} is used. This demonstrates the efficacy of our framework in mitigating interference between knowledge, which is one of the major problems of MTL. 

\subsection{Visualization of Knowledge Aggregation}\label{sec:visualization_knowledge_Aggregation}
\begin{figure}[t!]\small
    \begin{center}
    \begin{tabular}{@{}c c@{}}
    \multicolumn{2}{c}{\includegraphics[width=\columnwidth]{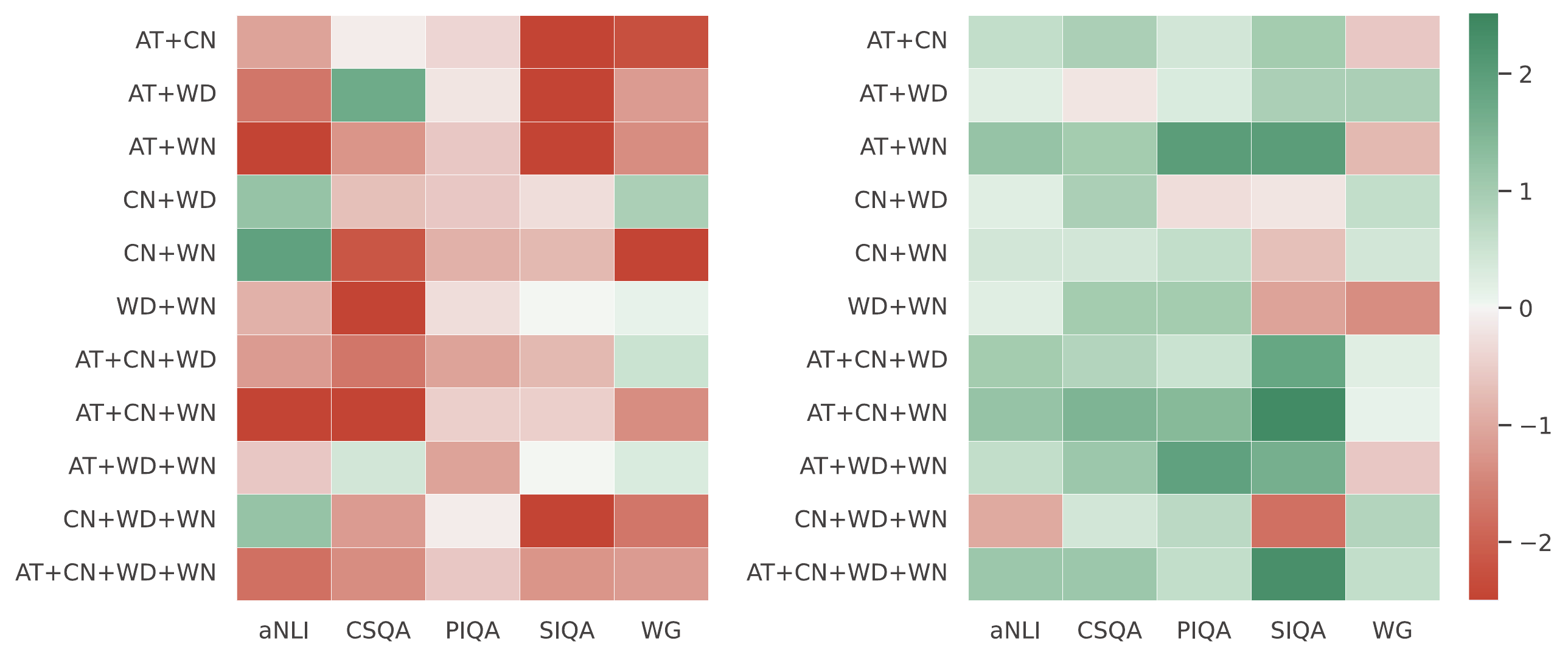}}\\

    \multirow{2}*{\qquad \qquad \quad (a) MTL \qquad }  & (b) zero-shot fusion \\ & w/ \emph{KG-C adapter}\\
    \end{tabular}
    \end{center}
    \caption{
    Relative improvement upon the STL on five benchmarks. The x- and y-axis indicate the benchmark and the combination of the KGs, respectively. The value of each cell indicates the relative performance improvement of using multiple KGs over the highest performance among STLs. The green and red colors denote the improvement or decrease of relative performance, respectively.
    }
    \label{fig:knowledge-aggregation}
\end{figure}

To verify the ability of our model to aggregate different types of KGs, we compare the relative performance gains of MTL and zero-shot fusion with \emph{KG-C adapter} when increasing the number of KGs. The performance of all KG-combinations for each framework is presented in Table~\ref{table:mtl_result} and Table~\ref{table:zeroshot_fusion_kgc_result}. We visualize the improvement of performance for five benchmark development sets, leveraging heatmaps in Figure~\ref{fig:knowledge-aggregation}. Here, for the sake of brevity, we denote our framework with \emph{KG-C adapter} as our method. 

For MTL in Figure~\ref{fig:knowledge-aggregation} (a), the color of the cell denotes the relative improvement of MTL with the combination of KGs over the best performance among the STL-PLM of KGs. Also, for our method in Figure~\ref{fig:knowledge-aggregation} (b), the relative improvement is measured based on the best performance among the STL-Adapter of KGs, considering the difference of the base architecture for MTL (i.e. PLM) and zero-shot fusion (i.e. PLM with adapter). The green and red colors denote the increase and decrease of performance, respectively, when using multiple KGs together. The greener color on the cells indicates that the approach benefits from an increasing number of KGs, which implies aggregating knowledge successfully.

In Figure~\ref{fig:knowledge-aggregation}, while the MTL tends to show the decrease of the performance when more KGs are utilized for training, our method obtains relative performance improvement across most of benchmarks. In both framework, the slightly degraded performance of the combination of KGs without \texttt{ATOMIC} could be due to the strong alignment between \texttt{ATOMIC} and SIQA. Except for the above case, we can observe that as more KGs are leveraged, the color of the cell gets greener, which implies that our method gains more advantages for better performance. This demonstrates that our method enables knowledge aggregation for multiple KGs synergetically.

\section{Conclusion}

Despite the existence of various types of commonsense KGs, utilizing multiple KGs has not been explored enough in the commonsense reasoning field. Motivated by this, this paper proposes a modularized transfer learning framework to fuse the knowledge from multiple KGs efficiently for zero-shot commonsense reasoning. Our framework consists of KG modularization for \emph{expert adapter}, zero-shot fusion and \emph{KG-Classifier adapter}. Extensive experiments show that our framework obtains strong improvements over MTL on five commonsense reasoning benchmarks.

In the future, our work can be extended to adapt our methods to further various multiple KGs with studies of appropriate scale for KG modularization. In addition, based on our hypothesis that the existence of an optimal combination, we can explore the study for the optional use of modularized KG experts for the best transfer learning.

\section*{Acknowledgements}
This work was partly supported by Institute of Information \& communications Technology Planning \& Evaluation (IITP) grant funded by the Korea government (MSIT) (No.
2020-0-01361, Artificial Intelligence Graduate School Program (Yonsei University)) and (No. 2022-0-00077, AI Technology Development for Commonsense Extraction, Reasoning, and Inference from Heterogeneous Data) and the National Research Foundation of Korea (NRF) grant funded by the Korea government (MSIT) (No. 2021-11-1055). Jinyoung Yeo is a corresponding author.

\bibliographystyle{acl_natbib}
\bibliography{acl2022}

\clearpage
\appendix
\section{List of Notations}\label{ap:notation}
We summarize the notations used in this paper in Table~\ref{table:notation}.

\section{Synthetic QA}\label{ap:synthetic_qa}
We generate QA for four KGs (\texttt{ATOMIC}, \texttt{ConceptNet}, \texttt{WikiData} and \texttt{WordNet}) based on synthetic QA generation~\citep{ma2020knowledgedriven} without sample filtering. We use the prefixes for relation of triplet as shown in Table~\ref{ap:syntheticqa-prefix} for generating synthetic QA (refer to \citet{ma2020knowledgedriven}). Table~\ref{ap:syntheticqa-statistics} shows the statistics of the synthetic QA dataset from KGs. The samples of synthetic QA with source triplet are shown in Table~\ref{ap:1-syntheticqa-sample}.

\begin{table}[h!] \small
\centering
\renewcommand{\arraystretch}{1.1}
\begin{tabular}{ c c} 
\hline
\noalign{\hrule height 0.8pt}
relation & prefix\\
\noalign{\hrule height0.8pt}
    xAttr            & . PersonX is seen as \\
    xIntent          & . Before, PersonX wanted \\
    xNeed            & . Before, PersonX needed to \\
    xReact           & . As a result, PersonX felt \\
    xWant            & . As a result, PersonX wanted to \\ 
    xEffect          & . PersonX then\\
    oReact           & . As a result, others felt \\
    oWant            & . As a result, others wanted to\\ 
    oEffect          & . Others then\\
	Causes           & can cause [MASK]\\
	UsedFor          & can be used for [MASK]\\
	CapableOf        & is capable of [MASK]\\ 
	CausesDesire     & causes desire for [MASK]\\ 
	IsA.             & is a [MASK]\\
	SymbolOf         & is a symbol of [MASK]\\
	MadeOf           & can be made of [MASK]\\ 
	LocatedNear      & is often located near [MASK]\\
	Desires          & desires [MASK]\\
	AtLocation       & can be found at [MASK]\\
	HasProperty      & has property [MASK]\\
	PartOf           & is part of [MASK]\\
	HasFirstSubevent & starts by [MASK]\\
	HasLastSubevent  & ends by [MASK]\\
\hline
\noalign{\hrule height 0.8pt}
\end{tabular}
\caption{Prefixes used for synthetic QA dataset} 
\label{ap:syntheticqa-prefix}
\end{table}

\begin{table}[h] \small
\centering
\renewcommand{\arraystretch}{1.2}
\begin{tabular}{c c c c} 
\hline
\noalign{\hrule height 0.8pt}
KG & Train & Validation & Total \\
\hline
\noalign{\hrule height 0.8pt}
\texttt{ATOMIC}      & 534,833 & 60,289 & 595,122 \\
\texttt{ConceptNet}  & 363,645 & 19,140 & 382,785 \\
\texttt{WikiData}    & 42,342  & 2,229  & 44,571 \\
\texttt{WordNet}     & 256,922 & 13,523 & 270,445 \\
\hline
Whole                & 1,197,742 & 95,181 & 1,292,923 \\
\hline
\noalign{\hrule height 0.8pt}
\end{tabular}
\caption{Synthetic QA dataset statistics. Whole represents the combination of AT,CN,WD and WN.}
\label{ap:syntheticqa-statistics}
\end{table}

\begin{table}[t] \small
\renewcommand{\arraystretch}{1.2}
\begin{tabular}{l} 
\hline
\multicolumn{1}{p{0.45\textwidth}}{\raggedright \textbf{QA from} \texttt{ATOMIC}~\citep{sap2019atomic}} \\ \hline
\multicolumn{1}{p{0.45\textwidth}}{\raggedright ($e^h,r,e^t$): (Dana speeds on the highway., xAttr, risky)} \\ 
\multicolumn{1}{p{0.45\textwidth}}{\raggedright Q: Dana speeds on the highway. Dana is seen as} \\ 
\multicolumn{1}{p{0.45\textwidth}}{\raggedright A1: considerate \textbf{A2: risky}(\checkmark) A3: lazy} \\ 
\hline

\multicolumn{1}{p{0.45\textwidth}}{\raggedright \textbf{QA from} \texttt{ConceptNet}~\citep{speer2017conceptnet}} \\ \hline
\multicolumn{1}{p{0.45\textwidth}}{\raggedright ($e^h,r,e^t$): (pentode, IsA, vacuum tube)} \\ 
\multicolumn{1}{p{0.45\textwidth}}{\raggedright Q: pentode is a [MASK]} \\ 
\multicolumn{1}{p{0.45\textwidth}}{\raggedright A1: ascocarp A2: girls footwear \textbf{A3: vacuum tube}(\checkmark)} \\ 
\hline

\multicolumn{1}{p{0.45\textwidth}}{\raggedright \textbf{QA from} \texttt{WikiData}~\citep{wikidata}} \\ \hline
\multicolumn{1}{p{0.45\textwidth}}{\raggedright ($e^h,r,e^t$): (badminton, IsA, type of sport)} \\ 
\multicolumn{1}{p{0.45\textwidth}}{\raggedright Q: badminton is a [MASK]} \\ 
\multicolumn{1}{p{0.45\textwidth}}{\raggedright A1: fable  A2: juvenile justice \textbf{A3: type of sport}(\checkmark)} \\ 
\hline

\multicolumn{1}{p{0.45\textwidth}}{\raggedright \textbf{QA from} \texttt{WordNet}~\citep{miller1995wordnet}} \\ \hline
\multicolumn{1}{p{0.45\textwidth}}{\raggedright ($e^h,r,e^t$): (princewood, PartOf, genus Cordia)} \\ 
\multicolumn{1}{p{0.45\textwidth}}{\raggedright Q: princewood is part of [MASK]} \\ 
\multicolumn{1}{p{0.45\textwidth}}{\raggedright A1: shaddock \textbf{A2: genus Cordia}(\checkmark) A3: family Columbidae} \\ 
\hline
\end{tabular}
\caption{Synthetic QA examples. We use templates to convert a question ($e^{head},r$) into a natural language.} 
\label{ap:1-syntheticqa-sample}
\end{table}

\section{Commonsense Knowledge Graphs}\label{ap:commonsense_knowledge_graph}
A variety of KGs have been proposed to provide large-scale high quality collection of different commonsense knowledge types: \texttt{ATOMIC}~\cite{sap2019atomic} mainly consists of inferential knowledge based on \emph{if-then} relations (\eg, \emph{X repels Y's attack. As a result, Y wants to attack X again}). \texttt{ConceptNet}~\cite{speer2017conceptnet} focuses on the general knowledge including lexical and world knowledge using various relations between entities (\eg, \emph{isA}, \emph{PartOf} and \emph{UsedFor}). \texttt{WikiData}~\cite{wikidata} is a general KG which has a close relation with Wikipedia. \texttt{WordNet}~\cite{miller1995wordnet} is a large lexical source of words and taxonomical system.

\section{Dataset for Zero-shot Fusion} \label{ap:dataset_for_zeroshot_fusion}

For zero-shot fusion training, we use balanced mixture of synthetic QA from different KGs by random sampling. The statistics of dataset for zero-shot fusion is shown in Table~\ref{ap:fusion-statistics}. For validation dataset, we balance between the \texttt{ATOMIC}, \texttt{ConceptNet} and \texttt{WordNet} due to the lack of synthetic QA validation dataset from \texttt{WikiData}.

\begin{table}[h!] \small
\centering
\renewcommand{\arraystretch}{1.3}
\begin{tabular}{c c c c} 
\hline
\noalign{\hrule height 0.8pt}
KG & Train & Validation & Total \\
\hline
\noalign{\hrule height 0.8pt}
$+$\texttt{ATOMIC}      & 2,500 & 2,500 & 5,000 \\
$+$\texttt{ConceptNet}  & 2,500 & 2,500 & 5,000 \\
$+$\texttt{WikiData}    & 2,500 & 2,229 & 4,729 \\
$+$\texttt{WordNet}     & 2,500 & 2,500 & 5,000 \\
\hline
Total & 10,000 & 9,729 & 19,729 \\
\hline
\noalign{\hrule height 0.8pt}
\end{tabular}
\caption{Statistics of the dataset for zero-shot fusion}
\label{ap:fusion-statistics}
\end{table}

\begin{table*}[t!] \small
\centering
\renewcommand{\arraystretch}{1.2}
\begin{tabular}{l l} 
\hline
\noalign{\hrule height 0.8pt}
Notation & Meaning\\
\noalign{\hrule height0.8pt}
    $(e^{head}, r, e^{tail})$            & Triple of KG (head entity, relation, tail entity) \\
    $Q_i$            & Natural language Question of sample $i$ \\
    $A_i=\{A_{i,1},...,A_{i,m}\}$   & A set of answer options of sample $i$, $A_{i,j}$ denotes $j$-th answer option of sample $i (1 \leq j \leq m)$ \\
    $T_i=\{T_{i,1},...,T_{i,m}\}$ & Input sequences generated by concatenation of $Q_i$ and $A_i$ \\
    $w_t$           & A word $t$-th token in the sequence $T_{i,j}$ \\
    $label$          & the index of the correct answer for sample \\
    $\calD_{QA}^k$   & Synthetic QA generated by KG k, 1 $\leq$ $k$ $\leq$ $K$  \\
    $N_k$            & The number of samples for $\calD_{QA}^k$, 1 $\leq$ $k$ $\leq$ $K$ \\ 
    $\theta$            & Parameters for pre-trained LM \\ 
    $\Phi_{QA}^k$    & Parameters for the \emph{expert adapter} of KG $k$, 1 $\leq$ $k$ $\leq$ $K$  \\
    $\Phi_{KGC}$     & Parameters for the \emph{KG-Classifier adapter} \\
    $\Psi_{QA}$       & Parameters for the fusion layer \\
	$l$           & The index of Transformer layer \\
	$\textbf{W}_l^Q$          & Query matrix of fusion layer in $l$th Transformer layer \\
	$\textbf{W}_l^K$          & Key matrix of fusion layer in $l$th Transformer layer \\
	$\textbf{W}_l^V$          & Value matrix of fusion layer in $l$th Transformer layer \\
	$h_{PLM}^l$        & Hidden representation of PLM parameterized by $\theta$ in $l$th Transformer layer \\ 
	$h_{E}^{k,l}$        & Hidden representation of \emph{expert adapter} parameterized by $\Phi_{QA}^k$ in $l$th Transformer layer \\ 
	$h_{KGC}^l$        & Hidden representation of \emph{KG-Classifier adapter} parameterized by $\Phi_{KGC}$ in $l$th Transformer layer \\ 
\hline
\noalign{\hrule height 0.8pt}
\end{tabular}
\caption{Notations and their meanings} 
\label{table:notation}
\end{table*}

\section{KG-Classification Dataset}\label{ap:KG_classification_dataset}

We suggest KG-Classification dataset $\calD_{KGC}$ for \emph{KG-Classifier adapter} training. The example of transformation from synthetic QA dataset $\calD_{QA}$ is shown in Table~\ref{table:kg-classification-dataset}. The dataset size is equal to the whole dataset of synthetic QA (refer to Table~\ref{ap:syntheticqa-statistics}).

\begin{table}[h] \small
\renewcommand{\arraystretch}{1.2}
\begin{tabular}{l} 
\hline
\noalign{\hrule height 0.8pt}
\multicolumn{1}{p{0.45\textwidth}}{\raggedright \textbf{QA} $\rightarrow$ \textbf{KG-Classification} \texttt{ATOMIC}} \\ \hline
\multicolumn{1}{p{0.45\textwidth}}{\raggedright Q: Dana speeds on the highway. Dana is seen as} \\ 
\multicolumn{1}{p{0.45\textwidth}}{\raggedright A1: considerate \textbf{A2: risky}(\checkmark) A3: lazy} \\ 
\hline
\multicolumn{1}{p{0.45\textwidth}}{\raggedright S: Dana speeds on the highway. Dana is seen as risky. } \\ 
\multicolumn{1}{p{0.45\textwidth}}{\raggedright A: Atomic} \\ 
\hline
\noalign{\hrule height 0.8pt}
\multicolumn{1}{p{0.45\textwidth}}{\raggedright \textbf{QA} $\rightarrow$ \textbf{KG-Classification} \texttt{ConceptNet}} \\ \hline
\multicolumn{1}{p{0.45\textwidth}}{\raggedright Q: pentode is a [MASK]} \\ 
\multicolumn{1}{p{0.45\textwidth}}{\raggedright A1: ascocarp A2: girls footwear \textbf{A3: vacuum tube}(\checkmark)} \\ 
\hline
\multicolumn{1}{p{0.45\textwidth}}{\raggedright S: pentode is a vacuum tube. } \\ 
\multicolumn{1}{p{0.45\textwidth}}{\raggedright A: ConceptNet} \\ 
\hline
\noalign{\hrule height 0.8pt}
\multicolumn{1}{p{0.45\textwidth}}{\raggedright \textbf{QA}  $\rightarrow$ \textbf{KG-Classification} \texttt{WikiData}} \\ \hline
\multicolumn{1}{p{0.45\textwidth}}{\raggedright Q: badminton is a [MASK]} \\ 
\multicolumn{1}{p{0.45\textwidth}}{\raggedright A1: fable  A2: juvenile justice \textbf{A3: type of sport}(\checkmark)} \\ 
\hline
\multicolumn{1}{p{0.45\textwidth}}{\raggedright S: badminton is a type of sport. } \\ 
\multicolumn{1}{p{0.45\textwidth}}{\raggedright A: WikiData} \\ 
\hline
\noalign{\hrule height 0.8pt}
\multicolumn{1}{p{0.45\textwidth}}{\raggedright \textbf{QA} $\rightarrow$ \textbf{KG-Classification} \texttt{WordNet}} \\ \hline
\multicolumn{1}{p{0.45\textwidth}}{\raggedright Q: princewood is part of [MASK]} \\ 
\multicolumn{1}{p{0.45\textwidth}}{\raggedright A1: shaddock \textbf{A2: genus Cordia}(\checkmark) A3: family Columbidae} \\ 
\hline
\multicolumn{1}{p{0.45\textwidth}}{\raggedright S: princewood is part of genus Cordia. } \\ 
\multicolumn{1}{p{0.45\textwidth}}{\raggedright A: WordNet} \\ 
\hline
\noalign{\hrule height 0.8pt}
\end{tabular}
\caption{KG-Classification examples from synthetic QA dataset of each KG} 
\label{table:kg-classification-dataset}
\end{table}

\section{Zero-shot architecture with parameters}\label{ap:module_figure_parameter}
\begin{figure}[h]
\centering
    \includegraphics[width=\columnwidth]{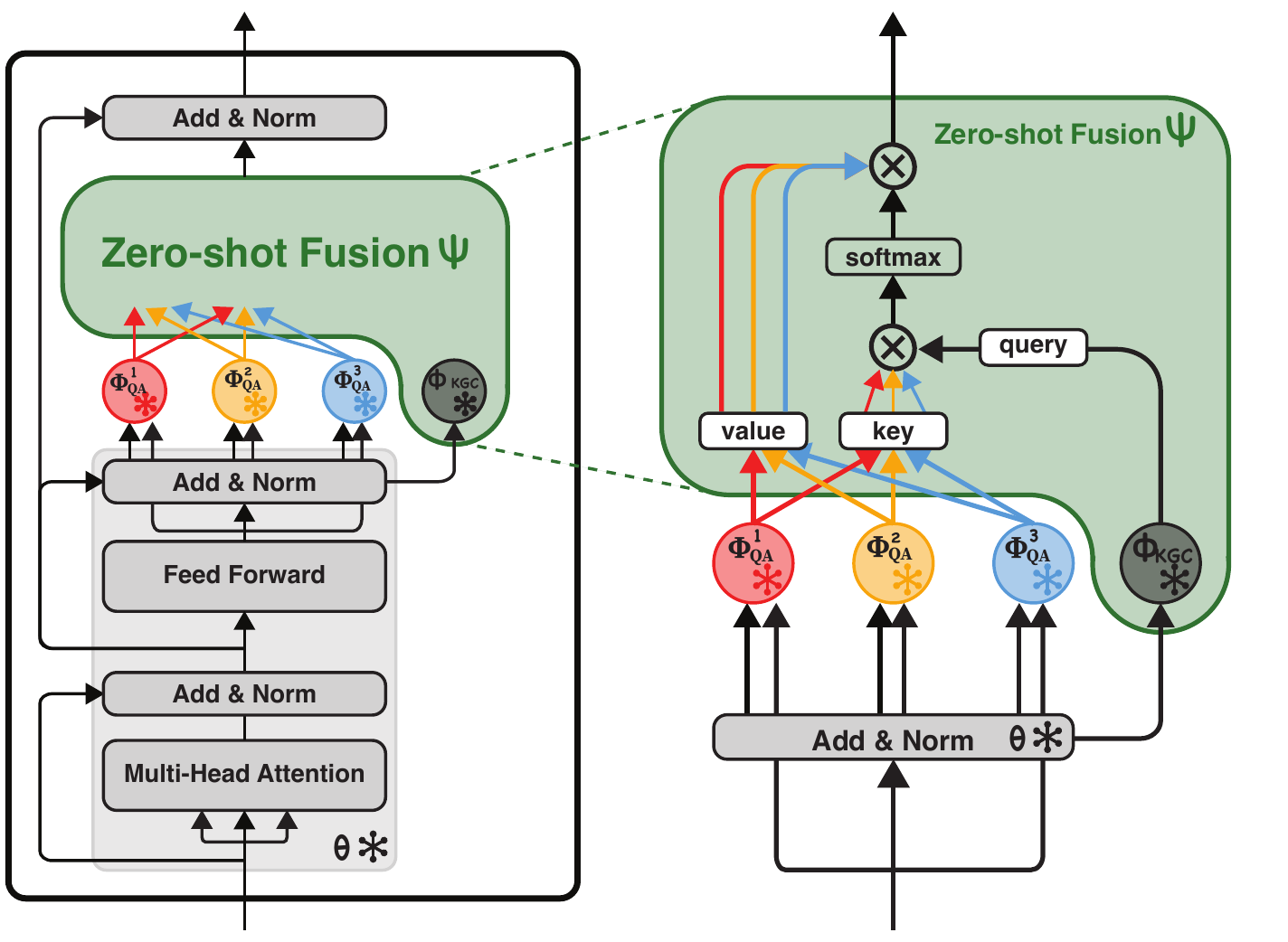}
\caption{Illustration of the zero-shot fusion architecture with parameters. Each colored circle represents \emph{expert adapters}, except the black circle which denotes \emph{KG-Classifier adapter}. $\ast$ indicates the fixed layer.}
\label{fig:module-figure-parameter}
\end{figure}

We describe the illustration of the zero-shot fusion architecture with parameters in Figure~\ref{fig:module-figure-parameter}.

\section{Commonsense Reasoning Benchmarks}\label{ap:commonsense_benchmarks}

\noindent \textbf{SocialIQA (SIQA)}~\citep{sap2019socialiqa} requires reasoning for emotional and social intelligence in everyday situations. Each QA consists of a context that comes from \texttt{ATOMIC}, a question which is based on the relations in \texttt{ATOMIC}, and 3 answer candidates. It contains 38,000 multiple-choice questions, which is generated by crowdsourcing.

\noindent \textbf{CommonsenseQA (CSQA)}~\citep{talmor2018commonsenseqa} evaluates a broad range of concept-level commonsense reasoning. Each multiple-choice question, answer and distractors are designed by crowdsourcing based on the \texttt{ConceptNet}.

\noindent \textbf{Abductive NLI (a-NLI)}~\citep{bhagavatula2019abductive} asks to infer the most plausible explanation based on the given causal situation to test abductive reasoning in narratives. Each sample consists of the beginning and the end of the story with two possible options to be an explanation for the given situation.

\noindent \textbf{PhysicalIQA (PIQA)}~\citep{bisk2020piqa} requires physical commonsense reasoning to select the most sensible solution for the given goal among the two choices. Its dataset is comprised of over 16,000 training samples, 2K validation samples, and 3K test samples.

\noindent \textbf{HellaSWAG (HSWAG)}~\citep{zellers2019hellaswag} is an evolved version of SWAG~\citep{zellers2018swag}, which asks to infer the most proper story based on the given situation. The dataset consists of 70K questions with four answer options.

\section{Implementation Details}\label{ap:implementation_details}
In all our experiments, we use max sequence length 128, batch size 32, weight decay 0.01, adam $\beta_1$ 0.9, adam $\beta_2$ 0.99, adam epsilion $1e^{-8}$, warm-up proportion 0.05, and margin 1.0. The experiments are conducted split across NVIDIA GeForce 3090 and NVIDIA RTX A5000.

\subsection{Baselines} The baseline models for STL-PLM and MTL are trained with learning rate $1e^{-5}$ for single epoch.

\subsection{Adapter} For \emph{expert adapters}, we use learning rate $8e^{-5}$ after tuning in $\{5e^{-6}, 8e^{-6}, 1e^{-5}, 5e^{-5}, 8e^{-5}, 1e^{-4}\}$. For \emph{KG-Classifier adapter}, we use learning rate $1e^{-5}$, batch size 64 for five epochs.

\subsection{Zero-shot fusion} After experiment with learning rates $\{1e^{-5}, 8e^{-5}\}$, we empirically find that a learning rate of $1e^{-5}$ works well on zero-shot fusion without/with \emph{KG-Classifier adapter}, respectively. Here, we set the attention drop probability 0.1. As we used extremely smaller subset of the synthetic QA dataset, zero-shot fusions are trained for five epochs. 

\section{Knowledge aggregation of zero-shot fusion}\label{ap:zeroshot_fusion_wo_kgc_relative_performance}
In order to validate the efficacy on knowledge aggregation of zero-shot fusion over the STL, we present the results of each framework with various combination of KGs in Table~\ref{table:mtl_result} and Table~\ref{table:zeroshot_fusion_kgc_result}.

\begin{table*}[t!]\small
\renewcommand{\arraystretch}{1.15}
\centering
\begin{tabular}{l l c c c c c c}
 \hline
 \noalign{\hrule height 0.8pt}
    \textbf{Model} & \textbf{KG} & \textbf{a-NLI}  & \textbf{CSQA} & \textbf{PIQA} & \textbf{SIQA} & \textbf{WG} & \textbf{Avg.}\\
\hline
\noalign{\hrule height0.8pt}
    \multirow{4}{*}{STL-PLM} &   AT  & 71.6	& 64.0 & 72.2	& 63.2 & 60.5 & 66.3 \\
                             &   CN  & 67.9	& 68.5 & 72.6	& 54.6 & 58.6 & 64.4 \\
                             &   WD  & 68.4	& 64.7 & 72.0	& 53.7 & 58.6 & 63.5 \\
                             &   WN  & 67.2	& 61.4 & 71.7	& 53.5 & 58.9 & 62.5 \\    
\hline
    \multirow{6}{*}{MTL}    & AT, CN & 70.5	& 68.4 & 72.2	& 60.1 & 58.2 & 65.9 \\
                            & AT, WD & 69.9	& 66.4 & 72.0	& 60.1 & 59.3 &	65.5 \\
                            & AT, WN & 69.1	& 62.7 & 71.6	& 59.1 & 59.1 &	64.3 \\
                            & CN, WD & 69.6	& 67.8 & 72.0	& 54.3 & 59.5 &	64.6 \\
                            & CN, WN & 69.8	& 66.3 & 71.7	& 53.8 & 56.4 &	63.6 \\
                            & WD, WN & 67.5	& 62.0 & 71.7	& 53.7 & 59.0 &	62.8 \\
\hline
    \multirow{4}{*}{MTL}    & AT, CN, WD & 70.4	& 66.8	& 71.5	& 62.4 & 61.0 & 66.4 \\
                            & AT, CN, WN & 68.5	& 65.7	& 72.1	& 62.7 & 59.1 & 65.6 \\
                            & AT, WD, WN & 71.0	& 65.1	& 71.1	& 63.2 & 60.8 & 66.2 \\
                            & CN, WD, WN & 69.6	& 67.3	& 72.5	& 52.0 & 57.2 & 63.7 \\
\hline
        MTL                 & AT, CN, WD, WN & 69.8 & 67.1 &	72.0& 61.9	& 59.3	& 66.0 \\
\hline
 \noalign{\hrule height 0.8pt}
 \end{tabular}
 
\caption{STL-PLM and MTL performance across five commonsense tasks in various combination of KGs. AT, CN, WD and WN represent \texttt{ATOMIC}, \texttt{ConceptNet}, \texttt{WikiData} and \texttt{WordNet}, respectively. We run our experiment with seed 42.
} 
\label{table:mtl_result}
\end{table*}

\begin{table*}[t!]\small
\renewcommand{\arraystretch}{1.15}
\centering

\begin{tabular}{l l c c c c c c}
 \hline
 \noalign{\hrule height 0.8pt}
    \textbf{Model} & \textbf{KG} & \textbf{a-NLI}  & \textbf{CSQA} & \textbf{PIQA} & \textbf{SIQA} & \textbf{WG} & \textbf{Avg.}\\
\hline
\noalign{\hrule height0.8pt}
    \multirow{4}{*}{STL-Adapter} &   AT  & 71.3	& 66.5	& 71.1 & 64.4 & 60.3 & 66.7 \\
                             &   CN  & 70.6	& 67.2	& 72.4 & 55.5 & 58.7 & 64.9 \\
                             &   WD  & 66.8	& 61.6	& 69.9 & 51.8 & 58.5 & 61.7 \\
                             &   WN  & 67.6	& 60.0  & 70.3 & 54.0 & 57.0 & 61.8 \\    
                             & AT,CN,WD,WN & 71.5	& 66.7 & 72.1 &	64.7	& 59.0 & 66.8 \\
\hline
    \multirow{6}{*}{zero-shot fusion w/\emph{KGC-adapter}}    & AT, CN & 71.9	& 68.1	& 72.8	& 65.4	& 59.7 & 67.6 \\
                            & AT, WD & 71.5	& 66.3	& 71.4	& 65.3	& 61.2 & 67.1 \\
                            & AT, WN & 72.5	& 67.5	& 73.1	& 66.4	& 59.5 & 67.8 \\
                            & CN, WD & 70.8	& 68.1	& 72.1	& 55.3	& 59.3 & 65.1 \\
                            & CN, WN & 71.0	& 67.6	& 73.0	& 54.8	& 59.1 & 65.1 \\
                            & WD, WN & 67.8	& 62.6	& 71.3	& 52.9	& 57.1 & 62.3 \\
\hline
    \multirow{4}{*}{zero-shot fusion w/\emph{KGC-adapter}}    & AT, CN, WD & 72.3	& 68.0	& 72.9	& 66.2	& 60.5	& 68.0 \\
                            & AT, CN, WN & 72.5	& 68.7	& 73.8	& 66.8	& 60.4	& 68.4 \\
                            & AT, WD, WN & 71.9	& 67.6	& 73.0	& 66.0	& 59.7	& 67.6 \\
                            & CN, WD, WN & 69.6	& 67.6	& 73.1	& 53.7	& 59.5	& 64.7 \\
\hline
\noalign{\hrule height0.8pt}
        zero-shot fusion w/\emph{KGC-adapter}                 & AT, CN, WD, WN & 72.4	& 68.3 & 73.0 &	66.7 & 60.9	& 68.3 \\
\hline
 \noalign{\hrule height 0.8pt}
 \end{tabular}
\caption{STL-Adapter and zero-shot fusion w/ \emph{KG-C adapter} performance across five commonsense tasks in various combination of KGs. AT, CN, WD and WN represent \texttt{ATOMIC}, \texttt{ConceptNet}, \texttt{WikiData} and \texttt{WordNet}, respectively. Whole represents the combination of AT, CN, WD and WN. We run our experiment with seed 42.
} 
\label{table:zeroshot_fusion_kgc_result}
\end{table*}

\begin{algorithm*}[h!]
\textbf{Input:} PLM parameters $\theta$, $K$ KGs \\
\textbf{Output:} Reasoning model parameters $(\theta, \{\Phi_{QA}^k\}_{k=1}^K, \Phi_{KGC}, \Psi_{QA})$ \\
$\{\calD^k_{QA}\}_{k=1}^K\leftarrow$ Generate synthetic QA samples from multiple KGs (Eq.~\ref{eq:qadata})\\
$\calD_{KGC} \leftarrow$ Generate KG classification samples from multiple KGs (Eq.~\ref{eq:kgcdata})\\
\For{each KG $k=1,...,K$}{
    $\Phi_{QA}^k \leftarrow \operatorname*{argmin}_{\Phi} 
    \calL_{QA}(\mathcal{D}^{k}_{QA} ; \theta, \Phi)$  (Eq.~\ref{eq:qamodel})
}
$\Phi_{KGC} \leftarrow \operatorname*{argmin}_{\Phi} \sum_{i=1}^{M} \calL_{KGC}(\calD_{KGC} ; \theta, \Phi)$ (Eq.~\ref{eq:kgcmodel})\\
$\Psi_{QA} \leftarrow \operatorname*{argmin}_{\Psi} \sum_{k=1}^{K} \calL_{QA}(\mathcal{D}^k_{QA}; \theta, \{ \Phi_{QA}^k\}_{k=1}^K, \Phi_{KGC}, \Psi)$ (Eq.~\ref{eq:fusionmodel} and~\ref{eqn:mean_query})\\
\textbf{return}~ $(\theta, \{\Phi_{QA}^k\}_{k=1}^K, \Phi_{KGC}, \Psi_{QA})$
\caption{Proposed framework for zero-shot commonsense reasoning}\label{alg:ourframework}
\end{algorithm*}

\end{document}